\documentclass[conference]{IEEEtran}  % Comment this line out if you need a4paper
\usepackage{times}

\IEEEoverridecommandlockouts                              % This command is only needed if 
\usepackage[numbers]{natbib}
\usepackage{multicol}
\usepackage[bookmarks=true]{hyperref}
\usepackage[font=small,labelfont=bf]{caption}
\usepackage{hyperref}
\usepackage{graphicx}
\usepackage{prettyref}
\usepackage{algorithm}
\usepackage{algpseudocode}
\usepackage{amsmath}
\usepackage{amssymb}
\usepackage{booktabs}
\usepackage{mathtools}
\usepackage{xcolor}

\algnewcommand{\LineComment}[1]{\State \(\triangleright\) #1}
\algdef{SE}[DOWHILE]{Do}{doWhile}{\algorithmicdo}[1]{\algorithmicwhile\ #1}
\newcommand*{\colorboxed}{}
\def\colorboxed#1#{%
  \colorboxedAux{#1}%
}
\newcommand*{\colorboxedAux}[3]{%
  \begingroup
    \colorlet{cb@saved}{.}%
    \color#1{#2}%
    \boxed{%
      \color{cb@saved}%
      #3%
    }%
  \endgroup
}

\newrefformat{Fig}{Figure~\ref{#1}}
\newrefformat{fig}{Figure~\ref{#1}}
\newrefformat{par}{Section~\ref{#1}}
\newrefformat{appen}{Appendix~\ref{#1}}
\newrefformat{sec}{Section~\ref{#1}}
\newrefformat{sub}{Section~\ref{#1}}
\newrefformat{table}{Table~\ref{#1}}
\newrefformat{alg}{Algorithm~\ref{#1}}
\newrefformat{Alg}{Algorithm~\ref{#1}}
\newrefformat{Def}{Definition~\ref{#1}}
\newrefformat{Thm}{Theorem~\ref{#1}}
\newrefformat{step}{Step~\ref{#1}}
\newrefformat{ln}{Line~\ref{#1}}
\newrefformat{eq}{Equation~\ref{#1}}
\newrefformat{pb}{Problem~\ref{#1}}
\newrefformat{it}{Item~\ref{#1}}
\newrefformat{te}{Term~\ref{#1}}
\def\Eqref Eq:#1:{\eqref{eq:#1}}
\newrefformat{Eq}{Equation~\Eqref#1:}

\newcommand{\E}[1]{\mathbf{#1}}
\newcommand{\TE}[1]{\textbf{#1}}

\newcommand{\argminP}[1]{\E{argmin}}
\newcommand{\argmax}[1]{\underset{#1}{\E{argmax}}}
\newcommand{\argmaxP}[1]{\E{argmax}}

\newcommand{\TWORCell}[2]{\begin{tabular}{@{}l@{}}#1 \\ #2\end{tabular}}

\newif\ifProofRead
\ProofReadfalse
\newif\ifArxiv
\Arxivfalse
\title{\Large Cloth Manipulation Using Random-Forest-Based Imitation Learning\vspace{-5px}}

\author{Biao Jia, Zherong Pan, Zhe Hu, Jia Pan, Dinesh Manocha \\% <-this % stops a space
\url{http://cs.unc.edu/~biao/robustm}
\vspace{-5px}
\thanks{Biao Jia is with the Department of Computer Science, University of Maryland at College Park. E-mail: {\tt\small biao@cs.umd.edu}.}
\thanks{Zherong Pan is with the Department of Computer Science, University of North Carolina at Chapel Hill. E-mail: {\tt\small zherong@cs.unc.edu}.}
\thanks{Zhe Hu is with the Department of Mechanical Biomedical Engineering, City University of Hong Kong.
 E-mail: {\tt\small zhe.hu@my.cityu.edu.hk}.}
\thanks{Jia Pan is with the Department of Computer Science, the University of Hong Kong. E-mail: {\tt\small jpan@cs.hku.hk}.}
\thanks{Dinesh Manocha is with Departments of Computer Science and Electrical \& Computer Engineering, University of Maryland at College Park. E-mail: {\tt\small dm@cs.umd.edu}.}
}
\begin{document}

\maketitle
\thispagestyle{empty}
\pagestyle{empty}

\maketitle

\begin{abstract}
We present a novel approach for manipulating high-DOF deformable objects such as cloth. Our approach uses a random-forest-based controller that maps the observed visual features of the cloth to an optimal control action of the manipulator. The topological structure of this random-forest is determined automatically based on the training data, which consists of visual features and control signals. The training data is constructed online using an imitation learning algorithm. We have evaluated our approach on different cloth manipulation benchmarks such as flattening, folding, and twisting. In all these tasks, we have observed convergent behavior for the random-forest. On convergence, the random-forest-based controller exhibits superior robustness to observation noise compared with other techniques such as convolutional neural networks and nearest neighbor searches.
\end{abstract}

\section{Introduction}\label{sec:intro}
High-DOF deformable object manipulation, such as cloth manipulation, is an important and challenging problem in robotics and related areas. It has many applications, including assisted human dressing~\cite{Clegg:2015:AHD:2809654.2766986}, cloth folding~\cite{Li:2015b}, sewing~\cite{6224880}, etc. Compared with rigid bodies or three-dimensional volumetric deformable objects \cite{1642072}, cloth can undergo large deformations and form wrinkles or folds, which greatly increases the complexity of cloth manipulation tasks. The possibility of such large deformations is the major challenge in designing a cloth manipulation controller.
\textcolor{black}{In a real-life cloth manipulation task, a typical robot only observes a single RGB(D) image of the cloth. 
As a result, we need robust methods that can perform such complex manipulation  tasks based on a single view observation. 
This involves inferring the 3D configuration of the cloth from the image-based representation and compute the appropriate control action. 
For example, if a robot manipulates a piece of cloth by holding two corners of the cloth mesh, then the controller should infer the desired end-effector positions of the robot.}%A well-designed controller should be able to represent different configurations of the cloth, including those with large deformations, from partial observations such as an RGB(-D) image. It should also map from the current observation to the optimal control signal.

\begin{figure}[ht]
\centering
\includegraphics[width=0.45\textwidth]{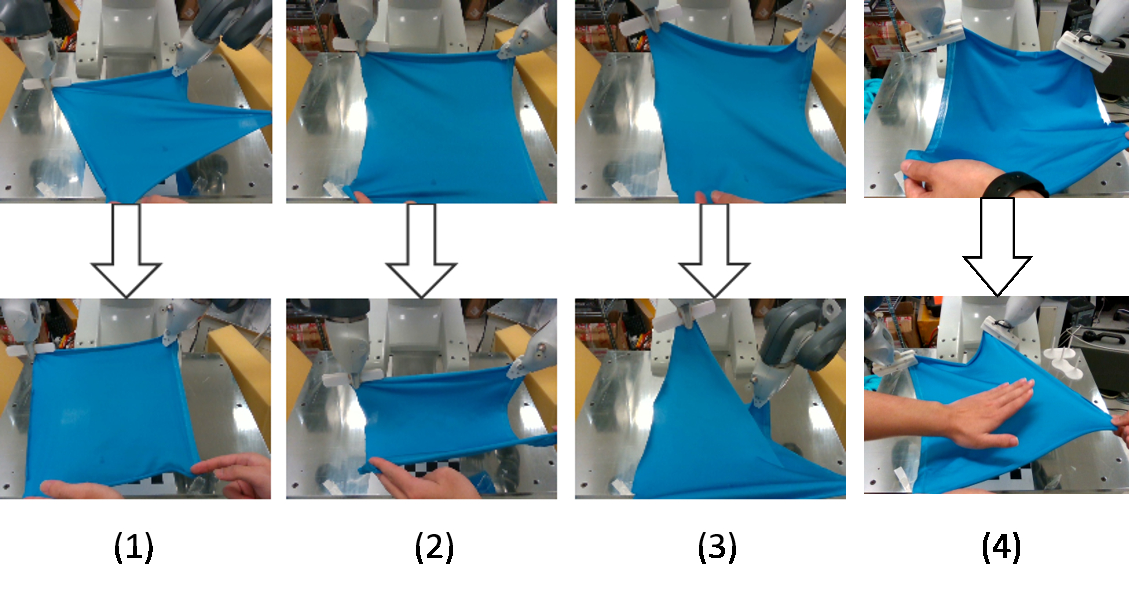}
\caption{ {\bf Manipulation Benchmarks:} We highlight the realtime performance of our algorithm on three basic robot-human collaboration tasks. (1): keep the cloth straight; (2): keep the cloth bent; (3): keep the cloth twisted.  \textcolor{black}{
(4): add noise to the human actions and the visual RGB-D outputs and evaluated the robustness of our approach. 
(5): evaluate the performance on complex tasks that simultaneously perform straightening, bending, and twist operations to highlight the benefits of our approach. 
}} %Our approach solves different tasks without manual parameter tuning.
\label{fig:real}
\vspace*{-5px}
\end{figure}
Several machine learning models have been proposed to parameterize such controllers, some of which have been used for cloth manipulations. Because of the recent development of deep (reinforcement) learning, one prominent method \cite{Levine:2016:ETD:2946645.2946684} is to represent feature extraction and controller parametrization as two neural networks, which are trained either jointly or separately. Other works, such as \cite{mnih2015humanlevel}, use one unified neural network architecture, but the structures of these neural networks are determined via trial and error. Recently, \cite{biao} represented the controller as a set of observations/control-signal pairs constructed manually. However, due to observation noise at runtime, it is not clear whether this constructed set can cover the experienced cases.

\TE{Main Result}: In this paper, we present a new method for cloth manipulation. Our method represents the controller as a random-forest. The random-forest takes the observation of the cloth configuration, an RGB(-D) image, as input. It then classifies the input by bringing it to a leaf-node of each decision tree. The optimal control signals are stored on the leaf-node and used as controller outputs. The random-forest is trained iteratively using imitation learning by collecting a dataset online. In each iteration, more data are collected and the random-forest is retrained to be more robust to observation noises.

Compared with other parametric models such as neural networks, random-forest is non-parametric and the number of leaf-nodes can be dynamically adjusted. As a result, arbitrarily complex cloth configurations can be represented as more training data are provided. Compared with other non-parametric methods such as nearest neighbor, random-forest exhibits better robustness in terms of avoiding over-fitting. We show that as more iterations of imitation learning are performed, the number of leaf-nodes in a random-forest will converge. 

We compare the performance of different controller models on three cloth manipulation tasks involving large deformations: cloth flattening, cloth folding, and cloth twisting. The results show that our model always outperforms nearest neighbor~\cite{biao} and neural networks in terms of matching optimal control signals and robustness to noise. In addition, the number of leaf-nodes converges as imitation learning progresses.

The rest of the paper is organized as follows. \prettyref{sec:related} reviews related works. In \prettyref{sec:prob}, we introduce the notation and formulate the problem. In \prettyref{sec:technique}, we provide details for training the random-forest-based controller. Finally, we highlight the performance on challenging benchmarks in \prettyref{sec:result} and compare the performance with prior methods.

\section{Related Work}
\label{sec:related}
In this section, we give a brief summary of prior works on large deformation and manipulation,  dimension reduction, and controller optimization.

\noindent \TE{Large Deformation and Manipulation}: Different techniques have been proposed for motion planning for deformable objects. Most of these works (e.g., \cite{1642072,6094946,4359263}) focus on volumetric objects such as a deforming ball or linear deformable objects such as steerable needles. By comparison, cloth-like thin-shell objects tend to exhibit more complex deformations, forming wrinkles and folds. Current solutions for thin-shelled manipulation problems are limited to specific tasks, including folding \cite{Li:2015b,yuba2017unfolding,stria2014garment}, ironing \cite{7487788}, sewing \cite{6224880}, and dressing \cite{Clegg:2015:AHD:2809654.2766986}. On the other hand, deformable body tracking solves a simpler problem, namely inferring the 3D configuration of a deformable object from sensing inputs. There is literature on deformable body tracking, which infers the 3D configuration from sensor data \cite{6630714,Wang:2015:DCM:2809654.2766911,10.1109/MCG.2015.96}. However, these methods usually require a template mesh as a priori and are mainly limited to handling small deformations.

\noindent \TE{Dimension Reduction}: Previous DOM methods use various feature extraction and dimensionality reduction techniques, including SIFT-features~\cite{7487788}, HOW-features \cite{biao}, and depth-based features~\cite{Doumanoglou2014,6906974,ramisa2013finddd}. Recently, deep neural networks have also been used as general-purpose feature extractors. They have also been used to manipulate low-DOF articulated bodies \cite{Levine:2016:ETD:2946645.2946684} and in DOM applications \cite{yang2017repeatable,tanaka2018emd}. For simplicity, our random-forest uses HOW-features as inputs. Another feature recently proposed in \cite{8258951} represents cloth using a small set of feature points. However, these feature points can only characterize small-scale deformations because there can be a lot of occlusions under large deformations. 

\noindent \TE{Controller Optimization}: In robotics, reinforcement learning \cite{Sutton:1998:IRL:551283}, imitation learning \cite{Hussein:2017:ILS:3071073.3054912}, and direct trajectory optimization \cite{Stengel:1986:SOC:26887} have been used to compute optimal control actions. Trajectory optimization, or a model-based controller, has been used in \cite{Li:2015b,7487788,6943185} for DOM applications. Although the resulting algorithms tend to be accurate, these methods cannot be used for realtime applications. For low-DOF robots such as articulated bodies \cite{6386109}, researchers have developed realtime trajectory optimization approaches, but it is difficult to extend them to deformable models due to the high simulation complexity of such models. Currently, realtime performance can only be achieved through learning-based controllers \cite{Doumanoglou2014,6906974,biao,yang2017repeatable}, which use supervised learning to train realtime controllers. However, as pointed out in \cite{AISTATS2011_RossGB11}, these methods are not robust in handling unseen data. Therefore, we further improve the robustness by using imitation learning. Apart from imitation learning used in this work, realtime cloth manipulation controllers can also be optimized using reinforcement learning methods as done in \cite{unknown,8345172,6697007}. Recently, \cite{7353475,Schulman2016,Lee2015LearningFM} proposed using non-rigid registration to transfer human demonstrations of cloth manipulations to real robots and \cite{7429768} required an adaptive cloth simulator to predict the future state of a cloth. However, these methods require the knowledge of full 3D cloth geometries, which are not available in our applications.
\begin{table}[t]
\setlength{\tabcolsep}{15pt}
\begin{tabular}{ll}
\hline  
Symbol & Meaning   \\
\hline  
$\mathcal{C}$ & 3D configuration space of the cloth  \\
$\E c$ & a configuration of the cloth   \\
$\mathcal{O}(\E c)$ & an observation of cloth  \\
$\E c^*$ & target configuration of the cloth  \\
$\E x$ & robot end-effectors' grasping points	\\
$\E x^*$ & optimal grasping points returned by the expert	\\
$P$ & transfer function encoding cloth dynamics	\\
$\E{dist}$ & distance measure between two observations	\\
\hline
$\pi$ & DOM-control policy	\\
$\alpha$ & random-forest topology	\\
$\beta$ & controller parameters	\\
$\gamma$ & confidence of leaf-node	\\
$\theta$ & parameter sparsity	\\
\hline
$K$ & the number of decision trees	\\
$l_k$ & a leaf-node of $k$-th decision tree	\\
$l_k(\mathcal{O}(\E c))$ & the leaf-node that $\mathcal{O}(\E c)$ belongs to \\
$\mathcal{L}$ & labeling function for optimal actions	\\
$\mathcal{F}$ & feature transformation for observation	\\
\hline
\end{tabular}
\caption{\label{Fig:param} Symbol table.}
\vspace{-5px}
\end{table}

\section{Problem Formulation}\label{sec:prob}
In this section, we introduce our notations and formulate the problem. Our goal is to compute a realtime feedback controller to deform a cloth into an unknown target configuration. We denote the 3D configuration space of the cloth as $\mathcal{C}$. Typically, a configuration $\E c \in\mathcal{C}$ can be discretely represented as a 3D mesh of cloth and the dimension of $\mathcal{C}$ can be in the thousands. However, we assume that only a partial observation $\mathcal{O}(\E c)$ is known, which is an RGB-D image from a single, fixed point of view in our case. The goal of the controller is to transform $\E c$ into a target configuration $\E　c^*$. We assume that, over the entire process of control, the robot grasps the cloth at a fixed set of $N$ points whose coordinates are $\E x$, where $|\E x|=3N$ and the control action is constituted by the desired positions of these grasping points, denoted as $\E x^*$. Therefore, the controller corresponds to a function:
\begin{eqnarray}
\label{eq:ctrl}
\E{x}^* = \pi(\mathcal{O}(\E{c})|\beta),
\end{eqnarray}
where $\beta$ are its learnable parameters. Given $x^*$, the corresponding joint angles of the robot can then be determined via conventional inverse kinematics. Given the control action, the configuration of the cloth and the grasping points can be given by the following distribution:
\begin{eqnarray}
p(\E c_{i+1},\E{x}_{i+1}|\E c_i,\pi(\mathcal{O}(\E c_i))).
\label{eq:p}
\end{eqnarray}
This distribution can be a cloth simulator \cite{Narain:2012:AAR:2366145.2366171} in a simulated environment or it can be obtained from a real-life robot. Note that, although the action is the desired grasping points ($\E{x}^*$), $\E{x}^*$ and $\E{x}_{i+1}$ are generally not the same because the controller's output can violate a robot's kinematic or dynamic constraints.

\subsection{Controller Optimization Problem}
Our main goal is to optimize the learnable parameters $\beta$ to optimize the performance of the controller, $\pi$. This controller optimization problem can take different forms depending on the available information about $\E{c}^*$. If $\mathcal{O}(\E{c}^*)$ is known, then we can define a reward function: $R(\E c)=-\E{dist}(\mathcal{O}(\E c),\mathcal{O}(\E c^*))$, where $\E{dist}$ can be any distance measure between RGB-D images. In this setting, we want to solve the following reinforcement learning problem:
\begin{eqnarray}
\argmax{\alpha,\beta}&&\E{E}_{\tau\sim\pi}{\left[\sum_i^\infty\gamma^iR(\E c_i)\right]}
\end{eqnarray}
where $\tau=(\E c_1,\E c_2,\cdots,\E c_\infty)$ is a trajectory sampled according to $\pi$, $\gamma$ is the discount factor, and the subscript figures denote the timesteps. Another widely used setting assumes that $\mathcal{O}(c^*)$ is unknown, but that an expert is available to provide an optimal control action $\pi^*(\mathcal{O}(\E c))$. The expert is a ground truth controller following the definition of \cite{Hussein:2017:ILS:3071073.3054912}. In this case, we want to solve the following imitation learning problem:
\begin{eqnarray}
\argmax{\alpha,\beta}&&\E{E}_{\tau\sim\pi}{\left[\sum_i^\infty\gamma^i\E{dist}(\pi^*(\mathcal{O}(\E c_i)),\pi(\mathcal{O}(\E c_i)))\right]}
\end{eqnarray}
This expert can be easily acquired in a typical human-robot collaboration task. Our method is based on the imitation learning formulation.

\begin{figure}[t]
\centering
\scalebox{0.8}{
\includegraphics[width=0.5\textwidth]{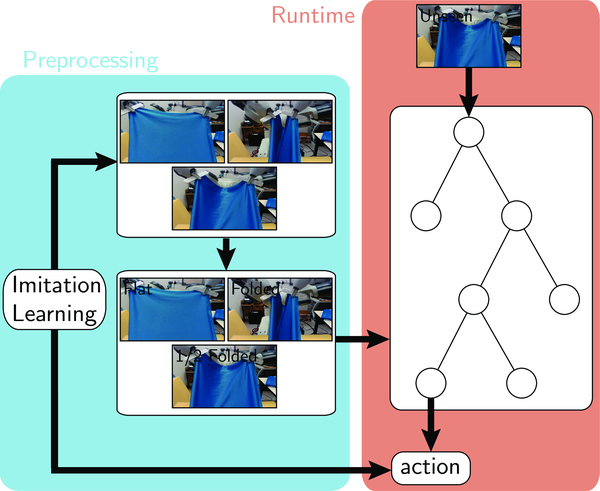}
\put(-130,110){(a)}
\put(-210,23){HOW Feature/Label (b)}
\put(-46,145){\parbox{.7in}{Random-Forest (c)}}
\put(-30,185){(d)}
\put(-50,8){(e)}
\put(-225,55){(f)}}
\caption{{\bf Approach Pipeline:} The pipeline of learning a random-forest-based DOM-controller that maps the visual feature (RGB-D image) to the control action. Given a sampled dataset (a), we first label each data point (shown as red text in (b)) to get a labeled dataset, (b). We then construct a random-forest to classify the images, (c). After training, the random-forest is used as  a controller. Given an unseen visual observation (d), the observation is brought through the random-forest to a set of leaf-nodes. The optimal control actions are defined on these leaf-nodes, (e). The entire process of labeling, classification, and controller optimization can be integrated into an IL algorithm, (f).}
\label{fig:framework}
\vspace{-5pt}
\end{figure}
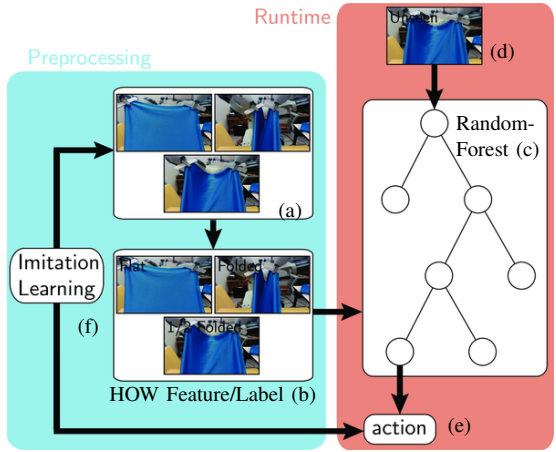
\section{Learning Random-Forest-Based Controller}\label{sec:technique}
To find the controller parameters, we use an imitation learning algorithm \cite{AISTATS2011_RossGB11}, which can be decomposed into two sub-steps: online dataset sampling and controller optimization. The first step samples a dataset $\mathcal{D}=\{\langle \mathcal{O}(\E{c}), \E{x}^* \rangle\}$, where each sample is a combination of cloth observation and optimal action. The second step optimizes the random-forest-based controller with respect to $\beta$, given $\mathcal{D}$.

\subsection{Feature Extraction}
Before constructing the random-forest from $\mathcal{D}$, we apply a feature transform to $\mathcal{D}$. Our raw observation of the cloth, $\mathcal{O}(\E{c})$, is an RGB-D image. it has been noted, (e.g., by \cite{4587617}) that applying a simple feature transform can improve the accuracy of a classifier such as random-forest. In addition, our input is a $320\times240$ RGB-D image of the cloth mesh, which corresponds to $76800$ entries each having three colors and one depth channel, which is high-dimensional. Therefore, a feature transform effectively reduces the dimensions of the input observation and makes the classifier more robust when the size of the dataset is small.

In our approach, we use HOW-features~\cite{biao} as the low-dimensional representation. HOW-features is a variant of HOG-features. HOW-features first applies Gabor filters to each patch of the image and then concatenates these patches, resulting in a $768$-dimensional feature space. Since each image patch is spatially localized, HOW-features requires each image to be aligned as a pre-processing step. Because our input is an RGB-D image, we can perform a foreground extraction using the depth-channel and then align the image to the center of the screen using the same procedure as in \cite{4587617}. We summarize this algorithm in \prettyref{Alg:Feature} and denote this feature transform as a function $\mathcal{F}$. The dataset after the feature transform is defined as $\bar{\mathcal{D}}=\{\langle \mathcal{F}\circ\mathcal{O}(\E{c}), \E{x}^* \rangle\}$. 
\setlength{\textfloatsep}{5pt}
\vspace{-5pt}
\begin{algorithm}[h]
\caption{\label{Alg:Feature} Feature extraction operation $\mathcal{F}$.}
\begin{algorithmic}[1]
\Require RGB-D image $\mathcal{O}(\E{c})$
\Ensure Extracted HOW-feature $\mathcal{F}\circ\mathcal{O}(\E{c})$
\State Foreground extraction using depth channel.
\State Resize/align image to the center of screen using \cite{4587617}.
\State Compute HOW-feature~\cite{biao}.
\end{algorithmic}
\end{algorithm}
\vspace{-10pt}

\subsection{Random-Forest Construction}
Our key contribution is to use a random-forest as the underlying learnable controller in an imitation learning framework. A random-forest is an ensemble of $K$ decision trees, where the $k$-th tree classifies $\mathcal{F}\circ\mathcal{O}(\E{c})$ by bringing it to a leaf-node $l_k(\mathcal{F}\circ\mathcal{O}(\E{c}))$, where $1\leq l_k(\mathcal{F}\circ\mathcal{O}(\E{c}))\leq L_k$ and $L_k$ is the number of leaf-nodes in the $k$-th decision tree. The random-forest makes its decision by classifying $\mathcal{F}\circ\mathcal{O}(\E{c})$ using every decision tree and then computing the average over all the decisions of the trees in the forest. To use an already constructed random-forest as a controller, we define an optimal control action $x_{l,k}^*$ so that the final action is determined by averaging:
\begin{eqnarray}
\label{eq:ctrl_rf}
x^* = \pi(\mathcal{O}(\E c)|\beta) = \frac{1}{K} \sum_{k=1}^K x_{l_k(\mathcal{F}\circ\mathcal{O}(\E c)),k}^*.
\end{eqnarray}
To construct the random-forest, we use a strategy similar to that in \cite{Shotton:2011:RHP:2191740.2192047}. We construct $K$ binary decision trees in a top-down manner, each using a random subset of $\mathcal{D}$. Specifically, for each node of a tree, a set of random partitions is computed and the one with the maximal Shannon information gain \cite{quinlan1986induction} is adopted. Each tree is grown until a maximum depth is reached or the best Shannon information gain is lower than a threshold. The optimal control action of a leaf-node is defined as the average of the control actions of the data sample belonging to that leaf-node.

\vspace{-5pt}
\subsection{Imitation Learning}
We use an imitation learning algorithm \cite{AISTATS2011_RossGB11} that includes two steps into an outer loop. \textcolor{black}{During each outer iteration, 
we query an expert, which in our case is a ground-truth hard-coded control algorithm. 
 Specifically, we generate a set of cloth simulation trajectories using a cloth simulator (\prettyref{eq:p}). 
 During each timestep of these trajectories, we query the expert to get an optimal control action $\pi^*(\mathcal{O}(\E c))$. This optimal control action is combined with the action proposed by our random-forest $\pi(\mathcal{O}(\E c))$. The combined action is fed to the simulator to get the next observation.} As a result, more data is added into $\mathcal{D}$ and a new random-forest, $\beta$, is constructed from a new $\mathcal{D}$. This algorithm is outlined in \prettyref{Alg:mainAlg}.

\vspace{-5px}
\setlength{\textfloatsep}{5pt}
\begin{algorithm}[h]
\caption{\label{Alg:mainAlg} Training DOM-controller using imitation learning algorithm.}
\begin{algorithmic}[1]
\Require Initial guess of $\beta$, optimal policy $\pi^*$
\Ensure Optimized $\beta$
\LineComment{imitation learning outer loop}
\While{imitation learning has not converged}
\label{ln:outer}
\LineComment{Generate training data based on current $\pi(\mathcal{O}(\E c)|\beta)$}
\State Sample $\mathcal{D}$ by querying $\pi^*$ as in \cite{AISTATS2011_RossGB11}
\LineComment{Extract HOW feature for each data sample}
\State Define $\bar{\mathcal{D}}=\emptyset$
\For{each $\mathcal{O}(\E c)$}
\State Extract HOW feature $\mathcal{F}\circ\mathcal{O}(\E c)$ as in~\cite{biao}
\State Define $\bar{\mathcal{D}}=\bar{\mathcal{D}}\bigcup\{\langle \mathcal{F}\circ\mathcal{O}(\E c),\pi^*(\mathcal{O}(\E c))\rangle\}$
\EndFor
\LineComment{Construct random-forest, i.e., $\beta$}
\For{$1\leq k\leq K$}
\State Sample random subset of $\bar{\mathcal{D}}$
\label{ln:rs}
\State Construct $k$-th binary decision tree using \cite{Shotton:2011:RHP:2191740.2192047}
\EndFor
\EndWhile
\end{algorithmic}
\end{algorithm}
\vspace{-15px}

\subsection{Analysis}
In typical DOM applications, data are collected using numerical simulations. Unfortunately, the high dimensionality of $\E{c}$ induces a high computational cost for simulations (i.e. evaluating $P$ in Equation \ref{eq:p}) and generating a large dataset can be quite difficult. Therefore, we design our method so that it can be used with a small number of data samples. Our method's performance relies on the random-forest's stopping criterion (i.e. the threshold of gain in Shannon entropy). We choose to use a large Shannon entropy threshold so that the random-forest construction stops early, leaving us with a relatively small number of leaf-nodes. We expect that, with a large enough number of imitation learning iterations, the number of nodes in each decision tree of the random-forest will converge. Indeed, such convergence can be guaranteed by the following Lemma. 

\TE{Lemma: }\textit{When the number of imitation learning iterations $N\to\infty$, the distribution incurred by the random-forest-based controller will converge to a stationary distribution and the expected classification error of the random-forest will converge to zero.}

\TE{Proof: } By assuming that \prettyref{Alg:mainAlg} generates a controller $\pi^n$ at the $n$-th iteration,  Lemma 4.1 of \cite{AISTATS2011_RossGB11} showed that $\pi^n$ incurs a distribution that converges when $n\to\infty$. Obviously, the number of data samples used to train the random-forest also increases to $\infty$ with $n\to\infty$. The expected error of the random-forest's classification on a stationary distribution converges to zero according to Theorem 5 of \cite{Biau:2012:ARF:2503308.2343682}. In \prettyref{sec:result}, we show that, empirically, the number of leaf-nodes in the random-forest also converges to a fixed value.

\section{Results}\label{sec:result}
We now describe our implementation and the experimental setup on both simulated environments and real robot hardware. We highlight the performance on several manipulation tasks performed by human-robot collaboration. We also highlight the benefits of using a random-forest-based controller by comparing our method with prior approaches. More implementation details are given in \cite{1802.09661}. 
\begin{figure}[ht]
\centering
\vspace*{-5px}
\includegraphics[width=0.3\textwidth]{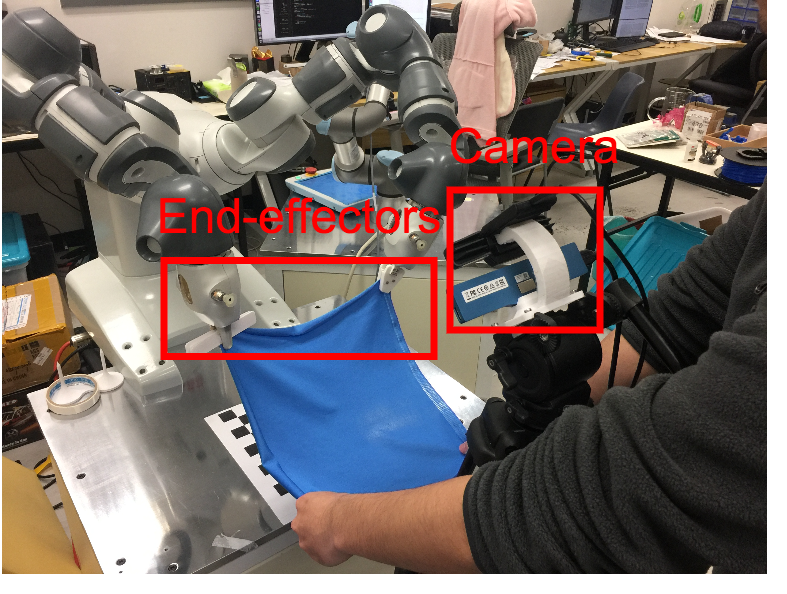}
\caption{{\bf Setup for Manipulation Tasks:} A dual-armed robot and a human are holding four corners of the cloth. We use a $12$-DOF dual-armed ABB YuMi and a RealSense RGB-D camera to perform complex manipulation tasks. Our goal is to manipulate a 35cm$\times$30cm rectangular-shaped piece of cloth. }
\label{fig:task}
\vspace{-15pt}
% \vspace{-5pt}
\end{figure}

\vspace{-5pt}
\subsection{Robot Setup}
We evaluate our method on a simulated environment. For the simulated environment, the robot's kinematics are simulated using Gazebo \cite{1389727} and the cloth dynamics are simulated using ArcSim \cite{Narain:2012:AAR:2366145.2366171}, a highly accurate cloth simulator. We use OpenGL to capture RGB-D in this simulated environment. Our goal is to manipulate a 35cm$\times$30cm rectangular piece of cloth with four corners initially located at: $v^0=(0,0,0), v^1=(0.3,0,0), v^2=(0,0.35,0), v^3=(0.3,0.35,0)$(m). Our manipulator holds the first two corners, $v^0,v^1$, of the cloth and the environmental uncertainty is modeled by having a human hold the last two corners, $v^2,v^3$, of the cloth so that we have $\E x\triangleq (v^0, v^1)^T$ and each control action is $6$-dimensional. The human could move $v^2,v^3$ to an arbitrary location under the following constraints:
\begin{eqnarray}
&&\|v^2-v^3\|\leq 0.3\text{m}	\\
&&\|(v^2,v^3)^T_{i+1}-(v^2,v^3)^T_i\|_\infty<0.1\text{(m/s)},
\end{eqnarray}
where the first constraint avoids tearing the cloth apart and the second constraint ensures that the speed of the human hand is slow. 
\begin{figure}[ht]
\centering
\vspace*{-10px}
\includegraphics[width=0.35\textwidth]{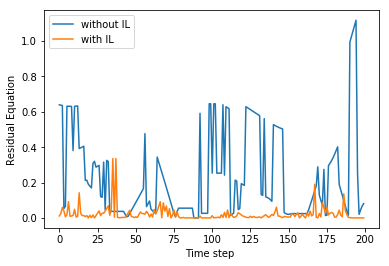}
\caption{{\bf Robustness of the imitation learning algorithm:} In a realtime human-robot interaction, we plot the mean action error (\prettyref{eq:err}). The blue curve shows the performance of a controller trained using only one imitation learning iteration (this choice corresponds to supervised learning~\cite{biao}) and the orange curve shows the performance of a controller trained with 20 iterations. We compare the residuals (\prettyref{eq:err}) between the two methods. Increasing the number of iterations in imitation learning significantly reduces the mean action error.}
\label{fig:robust}
\end{figure}

\subsection{Synthetic Benchmarks}\label{sec:sim}
To evaluate the robustness of our method, we design the 3 manipulation tasks listed below:
\begin{itemize}
\item Cloth should remain straight in the direction orthogonal to human hands. This is illustrated in \prettyref{fig:task} (a). Given $v^2,v^3$, the robot's end-effector should move to:
\begin{tiny}
\begin{align*}
v^0=v^2+0.35\frac{z\times(v^3-v^2)}{\|z\times(v^3-v^2)\|}\;
v^1=v^3+0.35\frac{z\times(v^3-v^2)}{\|z\times(v^3-v^2)\|}.
\end{align*}
\end{tiny}
\item Cloth should remain bent in the direction orthogonal to human hands. This is illustrated in \prettyref{fig:task} (b). Given $v^2,v^3$, the robot's end-effector should move to:
\begin{tiny}
\begin{align*}
v^0=v^2+0.175\frac{z\times(v^3-v^2)}{\|z\times(v^3-v^2)\|}\;
v^1=v^3+0.175\frac{z\times(v^3-v^2)}{\|z\times(v^3-v^2)\|}.
\end{align*}
\end{tiny}
\item Cloth should remain twisted along the direction orthogonal to human hands. This is illustrated in \prettyref{fig:task} (c). Given $v^2,v^3$, the robot's end-effector should move to:
\begin{tiny}
\begin{align*}
v^0=\frac{v^2+v^3}{2}+
0.31\frac{z\times(v^3-v^2)}{\|z\times(v^3-v^2)\|}+
0.15\frac{(v^3-v^2)\times(z\times(v^3-v^2))}{\|(v^3-v^2)\times(z\times(v^3-v^2))\|}\\
v^1=\frac{v^2+v^3}{2}+
0.31\frac{z\times(v^3-v^2)}{\|z\times(v^3-v^2)\|}-
0.15\frac{(v^3-v^2)\times(z\times(v^3-v^2))}{\|(v^3-v^2)\times(z\times(v^3-v^2))\|}.
\end{align*}
\end{tiny}
\end{itemize}
The above formula for determining $v^0,v^1$ is used to simulate an expert. \textcolor{black}{Note that these equations for the expert requires the knowledge of the four corner positions of the piece of cloth, 
and such information may not be available in a real robot system that only observes the cloth using a single RGB(D) image.
Therefore, we train our random-forest in a simulated environment.} These three equations assume that the expert knows the location of the human hands, but that robot does not have this information and it must infer this latent information from a single-view RGB-D image of the current cloth configuration. We also test the performance on complex benchmarks that combine flattening, folding, and twisting, or have considerable occlusion from a single camera.

\subsection{Transferring from Simulation to Real Robots} \label{sec:real}
\begin{figure*}[ht]
\centering
\includegraphics[width=\textwidth]{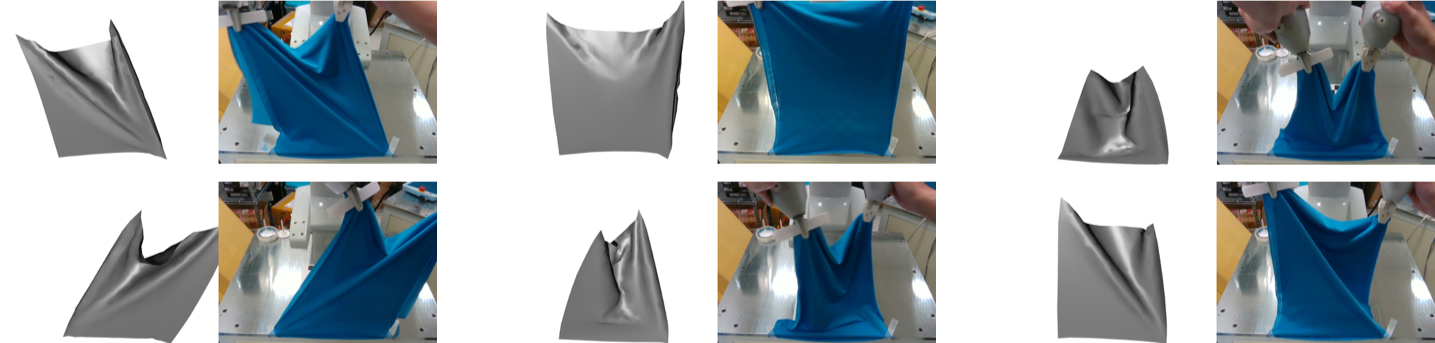}
\caption{{\bf Transferring from Simulation to Real Robots:} To transfer controller model, we trained on the cloth simulator to the real robot, we use camera calibration techniques. The figure demonstrates the result by changing the positions of end-effectors and fixing the other corners of the cloth. By applying a model trained with a calibrated cloth simulator to the real robot, we avoid having to train on the real robot.}
\label{fig:transform}
\end{figure*}
Although we have only evaluated our method on a simulated environment, we can also deploy our controller on real robot hardware. For the real robotic environment, we use a RealSense depth camera to capture 640$\times$480 RGB-D images and a 12-DOF ABB YuMi dual-armed manipulator to perform the actions, as illustrated in \prettyref{fig:task}.

To deploy our controller, we first use camera calibration techniques to get both the extrinsic and intrinsic matrix of the RealSense Camera. Second, we compute the camera position, camera orientation, and the clipping range of the simulator from the extracted parameters. Third, we generate a synthetic depth map using these parameters and train the three tasks using the random-forest-based controller parametrization and the imitation learning algorithm. \textcolor{black}{Finally, we randomly perturb the human hand positions when collecting training data to make our random-forest robust to observation noises. A similar technique is used in \cite{Tobin2017DomainRF}. We also add visual noises to the training samples and test the algorithm by posing objects between camera and object. } After that, we integrate the resulting controller with the ABB YuMi dual-armed robot and the RealSense camera via the ROS platform. As shown in \prettyref{fig:real}, with these identified parameters we can successfully perform the same tasks that were performed in the synthetic benchmarks on the real robot platform.

\begin{table}
\centering
\setlength{\tabcolsep}{3pt}
\begin{tabular}{ll}
\toprule
Name & Value \\
\midrule
Fraction term used in imitation learning algorithm \cite{AISTATS2011_RossGB11} & $0.8$	\\
Training data collected in each imitation learning iteration & $500$	\\
Resolution of RGB-D image & $640\times480$	\\
Dimension of HOW-feature used in \cite{biao} & $768$	\\
\TWORCell{Random-forest's stopping criterion when}{impurity decrease less than \cite{Shotton:2011:RHP:2191740.2192047}} & $1 \times 10^{-4}$	\\
\hline \\
\end{tabular}
\caption{\label{table:param} Meta-parameters used for training.}
\end{table}

\vspace{-5pt}
\subsection{Multi-task Controller}
 \textcolor{black}{Unlike single-task controller, a multi-task random-forest-based controller stores multiple actions in a leaf-node. Each observed image is classified by each decision tree in a manner that is similar to that of a single-task controller.
The leaf node chooses an action according to the id of the task.
In this benchmark, we train a 3-task controller for the 3 synthetic tasks in \prettyref{sec:sim}. And we transfer the controller to the real robot as benchmark (5) mentioned in \prettyref{fig:real}. 
We combines straightening, bending and twisting to show that our approach can perform complex tasks, as shown in the video. Moreover, we also show tasks which involve occlusion from a single camera viewpoint by adding noise to inputs.}

\begin{figure*}[t]
\vspace*{-10px}
\centering
\includegraphics[width=0.32\textwidth]{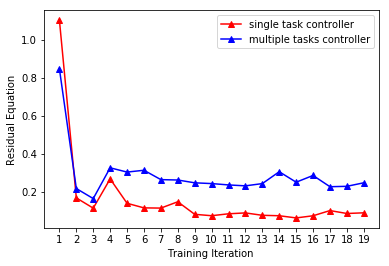}
\includegraphics[width=0.32\textwidth]{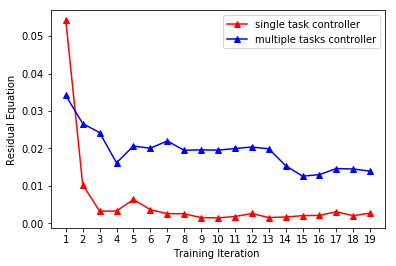}
\includegraphics[width=0.32\textwidth]{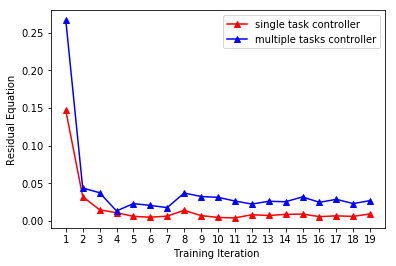}
\put(-420,68){(a)}
\put(-240,68){(b)}
\put(-80 ,68){(c)}
\caption{{\bf Multi-Task Controller vs. Single-Task Controller:} Residual (\prettyref{eq:err}) using a joint 3-task controller (blue) and a single-task controller (red). (a) Flatten the cloth; (b) Bend the cloth; (c) Twist the cloth. Both controllers converge after a few iterations of the imitation learning algorithm. The single-task controller performs slightly better than the multi-task controller with a relative action error of $0.4954$\%, but the difference is not significant.}
\label{fig:multi}
\vspace*{-15px}
\end{figure*}

We compare the performances of a single-task controller and a multi-task controller, both of which are based on random-forests. Again, during each evaluation in the simulated environment, the human hands move to $10$ random target positions $v^{2*},v^{3*}$. As shown in \prettyref{fig:multi} (red), we profile the residual (\prettyref{eq:err}). Our controller performs consistently well with a relative action error of $0.4954$\%. We then train a joint 3-task controller. This is performed by defining a single random-forest and defining $3$ optimal actions on each leaf-node. The performance of the 3-task controller is compared with that of the single-task controller in \prettyref{fig:multi}. The multi-task controller performs slightly worse in each task, but the difference is quite small.

\subsection{Complexity and Algorithm Properties}
As illustrated in \prettyref{Alg:mainAlg}, the complexity of our overall approach mainly depends on three parts: dataset sampling, feature extraction, and random-forest construction. When constructing a single decision tree based on the sampled dataset $\bar{D}$, the complexity has an upper bound of $O(|\bar{D}|^2)$. For the construction of a random-forest with $K$ decision trees, the complexity is $O(K|\bar{D}|^2)$.

\begin{figure*}[t]
\centering
\includegraphics[width=0.32\textwidth]{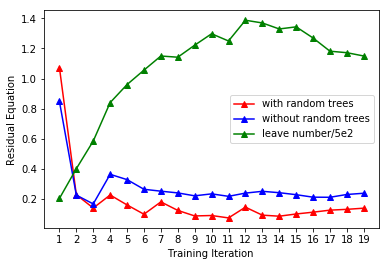}
\includegraphics[width=0.32\textwidth]{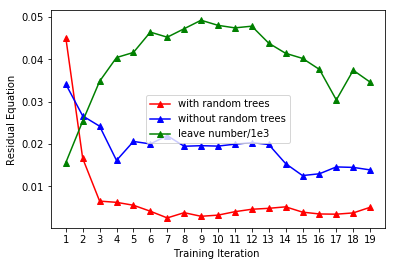}
\includegraphics[width=0.32\textwidth]{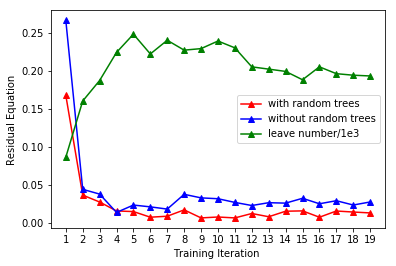}
\put(-420,78){(a)}
\put(-240,78){(b)}
\put(-80 ,78){(c)}
\caption{{\bf Controller with and without random-forest:} (Red): Residual (\prettyref{eq:err}) plotted against the imitation learning iterations (\prettyref{ln:outer} of \prettyref{Alg:mainAlg}). (Green): Number of leaf-nodes plotted against the imitation learning iterations. (Blue): Residual (\prettyref{eq:err}) plotted against the imitation learning iterations, which precludes random-forest construction. (a): Flatten the cloth; (b): Bend the cloth; (c): Twist the cloth.}
\label{fig:perf}
\end{figure*}

To evaluate the performance of each component in our method, we run several variants of \prettyref{Alg:mainAlg}. All the meta-parameters used for training are illustrated in \prettyref{table:param}. In our first set of experiments, we train a single-task random-forest-based controller for each task and profile the mean action error:
\begin{eqnarray}
\label{eq:err}
err&=&\sum_{\langle \mathcal{O}(c),\E x^* \rangle}\frac{1}{|\E x^*||\bar{\mathcal{D}}|}
\|\E x^*-\frac{1}{K}\E x_{l_k(\mathcal{F}\circ\mathcal{O}(c)),k}^*\|^2,
\end{eqnarray}
with respect to the number of imitation learning iterations (\prettyref{ln:outer} of \prettyref{Alg:mainAlg}). As illustrated in \prettyref{fig:perf} (red), the action error reduces quickly within the first few iterations and later converges. We also plot the number of leaf-nodes in our random-forest in \prettyref{fig:perf} (green). As more iterations are performed, the number of leaf-nodes in our random-forest also converges.

\subsection{Comparison With Other Solutions}
A key feature of our method is that it allows the robot to react to random human movements while the effect of these movements is indirectly reflected via a piece of cloth. This setting is similar to \cite{727484}. However, \cite{727484} assumes the 3D geometric mesh of cloth $c$ is known without any sensing error, which is not practical. 

Our method falls into a broader category of visual-servoing methods, but most previous work in this area (such as \cite{1703.11000}) has focused on navigation tasks and there is relatively little work on deformable body manipulation. \cite{7421959} based their servoing engine on histogram features, which is similar to our use of HOW-features. However, they use direct optimization to minimize the cost function ($\E{dist}(\mathcal{O}(\E c),\mathcal{O}(\E c^*))$), which is not possible in our case because our cost function is non-smooth in general.

Finally, our method is closely related to methods in \cite{Doumanoglou2014,6906974}, which also use random-forest and store actions on the forest. However, our method is different from prior methods in two ways. First, our controller is continuous in its parameters, which means it can be trained using an imitation learning algorithm. Moreover, we use both feature extraction and controller parametrization in the imitation learning algorithm \cite{AISTATS2011_RossGB11} so that both the feature extractor and the controller benefit from evolving training data.

To show the benefits of random-forest, we compare three different models of controllers: random-forest, linear regression, and neural network \cite{AISTATS2011_RossGB11}. During each evaluation in the simulated environment, the human hands move to $10$ random target positions $v^{2*},v^{3*}$. In \prettyref{table:cross}, we plot of the residual (\prettyref{eq:err}) of the tree methods against the number of imitation learning iterations. On the convergence of \prettyref{Alg:mainAlg}, the random-forest-based controller outperforms the two other opponents, exhibiting a lower residual.

To implement the neural-network-based controller, we use Tensorflow, which is a neural network toolkit. The structure of the neural network is fully connected and consists of a hidden layer of 128 neurons.
To implement the linear-regression-based controller, we use the apply the implementation from scikit-learn \cite{scikit-learn}, which is a standard machine learning toolkit. We use the standard parameters from the linear regression module.

\begin{table}
\vspace{-10px}
\centering
\setlength{\tabcolsep}{3pt}
\begin{tabular}{llllll}
\toprule
Training Set Proportion & $20\%$ & $40\%$ & $60\%$ & $80\%$ & $100\%$\\
\midrule
Random-Forest & $0.0154$ & $0.0078$ & $0.0046$ & $ 0.0040$ & $0.0038$	\\
Neural Network &  $0.0551$ & $0.0469$ & $0.0458$ & $0.0459$ & $0.0451$	\\
Linear Regression &  $1.66e18$ & $4.58e18$ & $8.77e17$ & $9.23e17$ & $8.82e-5$ \\
\hline \\
\end{tabular}
\vspace{-10px}
\caption{{\bf Comparison with Different Controllers:} Residual (\prettyref{eq:err}) of random-forest-based controller, neural-network-based controller \cite{AISTATS2011_RossGB11}, and linear regression controller, computed with different proportions of the training set. We use a dataset collected by an expert. The dataset contains $5702$ points and we randomly select $20\%$ of the data as the test dataset. 
The random-forest-based controller exhibits a lower residual. Linear regression increases residual on unseen data. A neural-network-based controller does not fit well when the size of the training set is limited. 
\label{table:cross}}
\end{table}

\vspace{-5pt}
\subsection{Benefits of Random-Forest}
% Comparing with the conventional regression based methods, our approach can get 
There are many standard techniques for computing low-dimensional controlling parameters from high-dimensional perceptual data such as RGB images and depth maps. These include standard regression models and neural-network-based models. We evaluate the performance of our algorithm along with the others.
The test involves measuring the residual of the manipulator as it moves towards the goal configuration based on the computed control parameters, as given by \prettyref{eq:err}.

We obtain best results in our benchmarks using a random-forest-based controller. Using the random-forest-based controller and the imitation framework requires fewer parameters to configure a task. Further, the computed control parameters are limited to the labels of the random-forest, which makes the controller robust to the unseen data. 
In practice, the random-forest-based imitation learning requires fewer computation resources which can enable the controller to be used in real-time applications. The performance is governed by the total number of iterations of the imitation learning. As the number of iterations of imitation learning grows, the residual \prettyref{eq:err} reduces. After reaching a certain iteration, the imitation learning contributes less to the performance enhancement. In other words, when the imitation learning framework converges, the overall performance of the controller is guaranteed.

\section{Conclusion, Limitations and Future Work}\label{sec:conclusion}
We present a novel controller parametrization for cloth manipulation applications. In our parametrization, the optimal control action is defined on the leaf-nodes of a random-forest. Further, both the random-forest construction and controller optimization are integrated with the imitation learning algorithm and evolve with training data. We evaluate our method using a 3-task cloth manipulation application. The result shows that our method can seamlessly handle feature extraction and controller parametrization problems. In addition, our method is robust to random noises in human motion and observations. Moreover, our controller parametrization can robustly adapt to evolving training data and quickly reduce the mean action error for real-time human robot interaction. 
During our evaluations, the controller performs consistently well in terms of accomplishing the cloth manipulation tasks, including the ones with very large cloth deformations. 
In terms of comparing with the traditional regression-based controller, our approach can model complex relationships between high dimensional input data and configurations of the controller.
Comparing with a neural-network-based controller, our approach can converge fast with limited input data, which makes it easier to adapt to unseen data.

One major limitation is that it is difficult to extend our method to reinforcement learning scenarios because our method is not differentiable when using a random-forest construction. Therefore, reinforcement learning algorithms such as the policy gradient method \cite{4058714} cannot be used. Another potential drawback is that our method is still sensitive to the random-forest's stopping criterion. In addition, we need additional dimension reduction, i.e. the HOW-feature,  and action labeling in the construction of the random-forest. In this work, labeling is done by mean-shift clustering of optimal actions, but in some applications where observations can be semantically labeled, it can be advantageous to label observations instead of actions. For example, in object grasping tasks, we can construct our random-forest to classify object types instead of classifying actions. \textcolor{black}{Finally, our method may not be suitable for high-level manipulation tasks such as cloth folding and laundry cleaning. These problems involve multiple smaller manipulation tasks which require a meta-algorithm that combines these tasks. In addition, these tasks usually require re-grasping between different stages of control, which is outside the domain of this paper.}

\section{Acknowledgement}
 This research is supported in part by ARO grant W911NF-19-1-0069, QNRF grant NPRP-5-995- 2-415, Intel, HKSAR General  Research  Fund (GRF) CityU 21203216, and NSFC/RGC Joint Research Scheme (CityU103/16-NSFC61631166002).

\begin{footnotesize}
\small{
\bibliographystyle{IEEEtran}
\bibliography{template}
}
\end{footnotesize}

\clearpage

\end{document}